\documentclass[10pt]{article} 
\usepackage[preprint]{tmlr}

\usepackage{amsmath,amsfonts,bm}

\def\eqref#1{equation~\ref{#1}}

\def\1{\bm{1}}

\DeclareMathAlphabet{\mathsfit}{\encodingdefault}{\sfdefault}{m}{sl}
\SetMathAlphabet{\mathsfit}{bold}{\encodingdefault}{\sfdefault}{bx}{n}

\usepackage{hyperref}
\usepackage{url}
\usepackage{makecell}
\usepackage{pifont}
\usepackage{graphicx}
\usepackage{amsmath,amsfonts,bm,amssymb,dsfont}

\usepackage[T1]{fontenc}
\usepackage{inconsolata} 
\usepackage{listings}
\usepackage{xcolor}

\definecolor{codegreen}{rgb}{0,0.6,0}
\definecolor{codegray}{rgb}{0.5,0.5,0.5}
\definecolor{codepurple}{rgb}{0.58,0,0.82}
\definecolor{backcolour}{rgb}{0.95,0.95,0.92}

\lstdefinestyle{mystyle}{
    backgroundcolor=\color{backcolour},   
    commentstyle=\color{codegreen},
    keywordstyle=\color{magenta},
    numberstyle=\tiny\color{codegray},
    stringstyle=\color{codepurple},
    basicstyle=\ttfamily\footnotesize,
    breakatwhitespace=false,         
    breaklines=true,                 
    captionpos=b,                    
    keepspaces=true,                 
    numbers=left,                    
    numbersep=5pt,                  
    showspaces=false,                
    showstringspaces=false,
    showtabs=false,                  
    tabsize=2
}

\title{Terra Nova: A Comprehensive Challenge Environment for Intelligent Agents}

\author{\name Trevor McInroe \email t.mcinroe@ed.ac.uk \\
    \addr The University of Edinburgh
}

\def\month{MM}  
\def\year{YYYY} 
\def\openreview{\url{https://openreview.net/forum?id=XXXX}} 

\begin{document}

\maketitle

\begin{figure}[!h]
    \centering
    \includegraphics[width=0.3\linewidth]{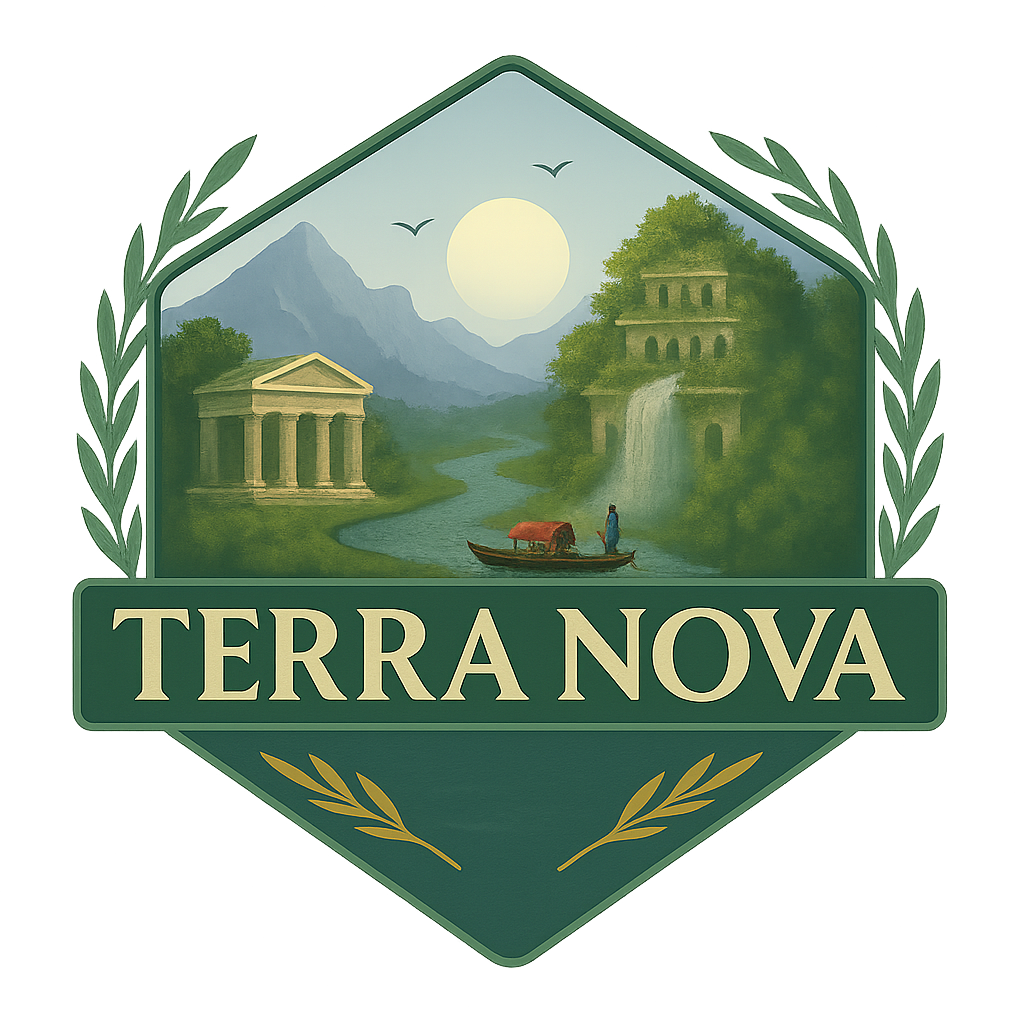}
\end{figure}

\section{Introduction}
We introduce Terra Nova\footnote{Codebase: \url{https://github.com/trevormcinroe/terra_nova/}}, a new \textit{comprehensive challenge environment} (CCE) for reinforcement learning (RL) research inspired by Civilization V~\citep{civ_v_game}. A CCE is a single environment in which multiple canonical RL challenges (e.g., partial observability, credit assignment, representation learning, enormous action spaces, etc.) arise simultaneously. Mastery therefore demands integrated, long-horizon understanding across many interacting variables. We emphasize that this definition excludes challenges that only aggregate unrelated tasks in independent, parallel streams (e.g., learning to play all Atari games at once). These aggregated multitask benchmarks primarily asses whether an agent can catalog and switch among unrelated policies rather than test an agent's ability to perform deep reasoning across many interacting challenges.

The purpose of CCEs is distinct from the purpose of many environments used in RL studies today. Today's environments generally attempt to isolate one specific challenge such that small, targeted studies on that challenge can occur fruitfully. We note that such environments are useful research tools. However, a CCE's purpose is to serve as a yardstick for progress and to highlight shortcomings of current general intelligence research. We assert that this research direction is important, as challenges rarely appear in isolation in real-world scenarios.

The use of CCEs in RL research has an influential history that led to significant advances. For example, research in StarCraft II~\citep{sc2le}, Dota 2~\citep{dota2}, and NetHack~\citep{nethack} has spurred innovation in search, planning, self-play, and other core areas of RL and control. CCEs are important for research because they expose the limitations of methods optimized for narrow tasks. As the field looks toward developing more capable and general agents, identifying new environments that meaningfully extend the CCE frontier, such as Terra Nova, is essential.

Terra Nova is inspired by Civilization V, a turn-based 4X strategy game\footnote{``4X'' stands for eXplore, eXpand, eXploit, eXterminate.} whose breadth of mechanics makes it a challenging testbed for general intelligence. Playing Terra Nova competently requires reasoning over a large set of diverse information streams and controlling hundreds of heterogeneous endpoints simultaneously. For example, agents must reason over a partially-observable map and disentangle multi-timescale credit assignments while searching vast hierarchical action-spaces to control units, cities, trade routes, diplomatic relations, and more. Perhaps most challenging of all, agents must continually assess a game state that includes five opponents to determine which of the many, mutually-exclusive victory paths they are most likely to achieve.

The remainder of this document proceeds as follows. First, we outline some of the challenges in Terra Nova and compare it with other current CCEs (\S\ref{sec:challenges}). Second, we briefly cover previous work in RL that has targeted aspects of Civilization as an environment (\S\ref{sec:previous-work}). Third, we formalize Terra Nova as a stochastic game (\S\ref{sec:formalizing}). Finally, we cover several key features of the Terra Nova software (\S\ref{sec:software}).

\begin{figure}[!t]
    \centering
    \includegraphics[width=0.95\linewidth]{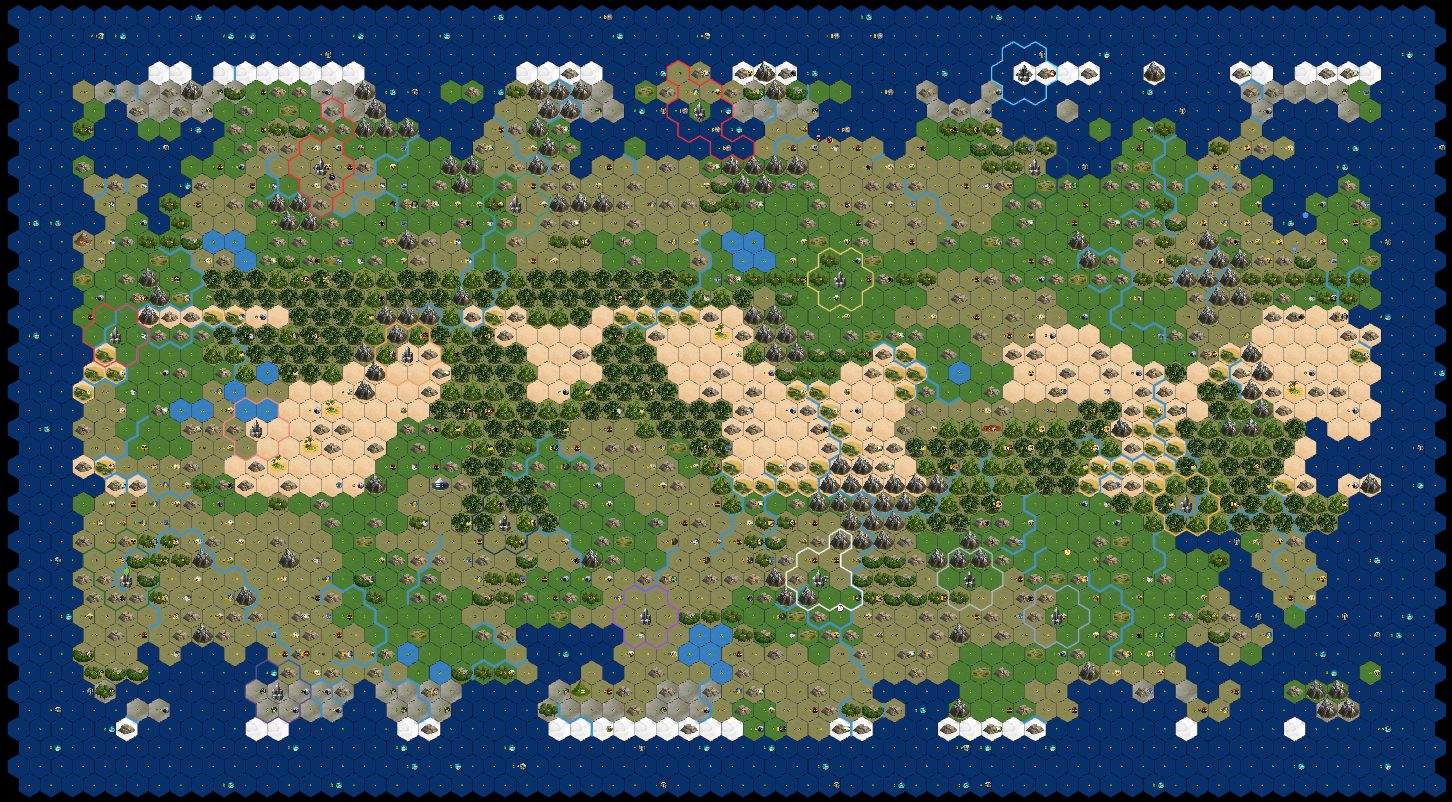}
    \caption{An example procedurally-generated Terra Nova map. The map is a central landmass surrounded by ocean and made of hexagonal tiles. The landmass is filled with various terrain types (e.g., desert, plains, grassland), features (e.g., oases, flood plains, jungles), elevation (e.g., flatland, hills, mountains), resources, water features, natural wonders, and more. For more information on maps, see the documentation here: \url{https://trevormcinroe.github.io/terra_nova_environment\#maps-mech}}
    \label{fig:map-zoomout}
\end{figure}

\section{Challenges in Terra Nova and Its Place Among Other CCEs}\label{sec:challenges}

Terra Nova provides a unique combination of challenges that sets it apart from other CCEs used previously. Also, Terra Nova's win mechanics differentiate it from all CCEs we examine. In Table~\ref{tab:comparison-table}, we compare Terra Nova with Starcraft 2~\citep{sc2le}, Dota 2~\citep{dota2}, Craftax~\citep{craftax}, NetHack~\citep{nethack}, NeuralMMO~\citep{neuralmmo}, and Diplomacy~\citep{diplomacy}. Below, we detail the characteristics shown in Table~\ref{tab:comparison-table} and give examples of how Terra Nova stands out. Then, we briefly describe additional challenges that agents face in a Terra Nova game.

\subsection{Comparing Terra Nova}

\begin{table}[t]
    \centering
    \begin{tabular}{>{\centering\arraybackslash}m{2cm} 
                    >{\centering\arraybackslash}m{2cm}
                    >{\centering\arraybackslash}m{2cm}
                    >{\centering\arraybackslash}m{2cm}
                    >{\centering\arraybackslash}m{2cm}
                    >{\centering\arraybackslash}m{2cm}}
        \makecell{Environment} & 
        \makecell{Opponent\\Structure} & 
        \makecell{Cooperation is\\Strategically\\Dominant} & 
        \makecell{Partial\\Observability} & 
        \makecell{Action\\Space} & 
        \makecell{Ways\\to Win} \\ \hline
        Starcraft 2 & 1v1 & \ding{55} & \ding{51} & Large & 1 \\
        Dota 2 & 1v1 & \ding{55} & \ding{51} & Large & 1 \\
        Craftax & Singleplayer & \ding{55} & \ding{51} & Small & 0 \\ 
        NetHack & Singleplayer & \ding{55} & \ding{51} & Small & 1 \\
        NeuralMMO & 1vMany & \ding{55} & \ding{51} & Small & 1 \\
        Diplomacy & 1vMany & \ding{51} & \ding{55} & Large & 1 \\
        Terra Nova & 1vMany & \ding{51} & \ding{51} & Large & 4
    \end{tabular}
    \caption{Comparison of CCEs across several dimensions. Elaboration on characteristics can be found in \S\ref{sec:challenges}.}
    \label{tab:comparison-table}
\end{table}

\textbf{Opponent structure} describes, in part, the competition dynamics. For example, ``1v1'' implies that either one agent/team wins or the other.  A ``singleplayer'' game is one that includes no opponents that can win the game. A ``1vMany'' game is a free-for-all game; i.e., more than two independent parties are trying to win the game. Terra Nova is a ``1vMany'' game.

\textbf{Cooperation is strategically dominant} indicates a game in which failing to cooperate with opponents places an agent at a significant disadvantage relative to those who do cooperate. Terra Nova's mechanics enable competing agents to cooperate on several facets. For example, agents can form trade deals, where resources, gold, or peace promises are exchanged. These trade deals are a core factor in growing an empire, and agents who do not trade quickly fall behind those who do.

\textbf{Partial observability} refers to settings in which global game state information is not always available to each agent. Terra Nova exhibits several forms of partial observability across different components of the game. For example, regions of the map that have never been explored are completely hidden. Once revealed, static features, such as terrain and resource locations, remain permanently visible. However, dynamic information, such as unit movement or newly founded cities, is only visible while the area is within the line of sight of the agent's units or cities. A different form of partial observability arises in the technology tree: while agents can infer which technologies have been unlocked globally, they cannot directly observe which specific agents have researched them.

\textbf{Action space} refers to the size and structure of the action space available to agents at each decision point. Several CCEs feature relatively small, flat action spaces. For example, Craftax has fewer than 50 unique actions. In contrast, Terra Nova presents an enormous ($\sim 10^{745}$)\footnote{Terra Nova's action space is factorized from roughly 450 sub-action spaces that range in size from two to 2772 options. For full details, see \url{https://trevormcinroe.github.io/terra_nova_documentation\#action-space}}, highly structured action space composed of many heterogeneous control surfaces. For example, on every turn, agents must allocate population within cities, select technologies from a branching technology tree, and issue movement and combat commands to units with distinct action spaces depending on unit types and game state context.

\textbf{Ways to win} indicates the number of unique win conditions. In contrast to other CCEs we consider, Terra Nova includes multiple mutually exclusive victory conditions such as military conquest, scientific advancement, cultural dominance, and diplomatic leadership. Progress toward one victory condition often directly undermines another. For instance, pursuing a cultural victory typically involves investing in civics and culture-related technologies at the expense of military or scientific development, making it harder to pivot to those alternative win paths later in the game.

\begin{figure}[ht]
    \centering
    \includegraphics[width=0.95\linewidth]{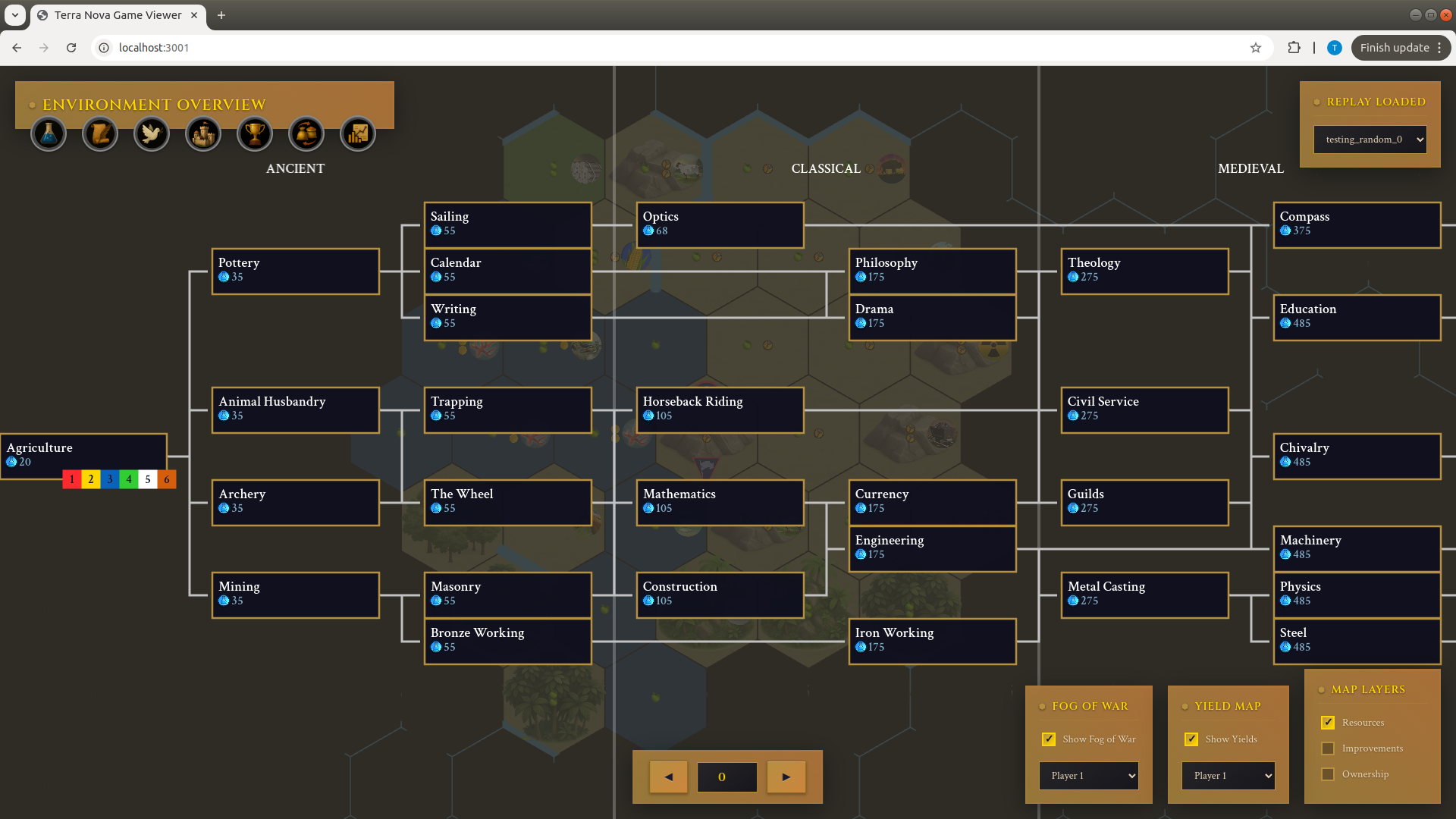}
    \caption{The beginning of the technology tree in Terra Nova. Specific technologies are represented with the rectangular emblems containing the technology name and science cost of unlocking. Prerequisite relationships are shown with the connecting gray lines. For example, to begin research on Engineering (towards the bottom of the 2nd column in the ``Classical'' Era), agents must first unlock Archery, Animal Husbandry, The Wheel, Mathematics, Mining, Masonry, and Construction.}
    \label{fig:viewer-tech-tree}
\end{figure}

\subsection{More Details on Terra Nova}
Terra Nova contains many challenges in addition to the ones mentioned above. Here we provide some detail on those challenges. Also, we provide more detail on the ``ways to win'' challenge described above that is outside of the scope of comparing to other environments.

\textbf{Observation space.} Terra Nova's observation space is large and structured. Each observation comprises over 100 elements. Elements are either scalars, vectors, arrays, or ``maps''. Map objects are arrays structured in the shape of the map and convey both information and physical location. Elements could be continuous values, categorical values, and could be marked as ``unknown'' due to the agent's limited knowledge of the game state.

\textbf{Multi-timescale credit assignment.} Actions in Terra Nova can provide benefits or costs across varying timescales. For example, improving a tile with a worker unit provides immediate benefits to the agent via increased yields, starting construction on a building could provide benefits in 20 turns when the construction completes, and the benefits of choosing a city-settling location might take hundreds of game turns to materialize.

\textbf{Theory of mind.} Choosing a strong strategy in Terra Nova requires forming accurate beliefs about opponent strategies. Forming these beliefs is made difficult due to imperfect information and partial observability.

\textbf{Opportunity cost.} Many mechanics in Terra Nova require investing multiple turns before events are triggered. For example, researching a technology could require dozens of turns before completion due to the tree-like structure of the technology pathways (i.e., before a technology can be chosen, prerequisite technologies in the tree must be completed). Due to this structure, choosing one path down the tree means purposefully forgoing progress down other pathways. See Figure~\ref{fig:viewer-tech-tree} for a depiction of the beginning of the technology tree.

\textbf{Generalization.} Map generation in Terra Nova is procedural and seed controlled. This allows for clean train/test split opportunities, meaning agents cannot simply memorize valuable action sequences.

\textbf{Ways to win.} Terra Nova has four distinct win types: (1) science, (2) domination, (3) cultural, and (4) diplomatic. We give detail on each win condition below.

\begin{figure}[!t]
    \centering
    \includegraphics[width=0.95\linewidth]{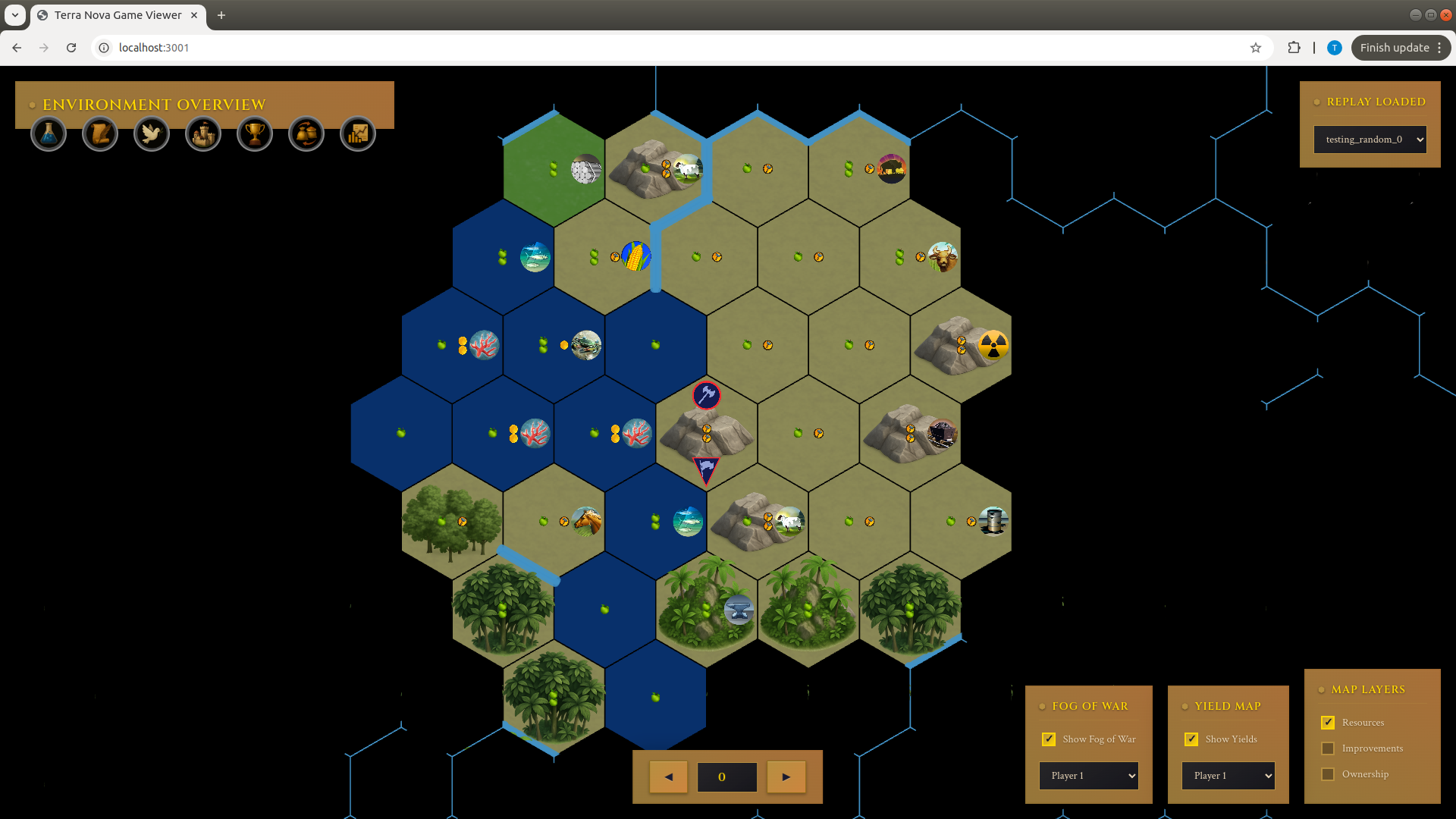}
    \caption{An example initial observation for an agent. In Terra Nova, agents begin the game with one Settler (represented with the triangle emblem and flag icon at the bottom of the center hex) and one Warrior (represented with the circle emblem and axe icon at the top of the center hex). This agent was initially spawned on the coast with a resource-rich section of the sea directly to its west and a large plains area to its east. On this initial turn, the agent must decide where to settle its capital city, weighing either settling in place on the coast or revealing more of the map (``unknown'' areas shaded in black) by moving its units.}
    \label{fig:map-first-turn}
\end{figure}

A science victory is achieved by constructing all components of the Space Shuttle: all three boosters, the engine, the cockpit, and the stasis chamber. Before construction on the parts can begin, agents must first complete construction of the Apollo Program and have researched the technology prerequisites of Rocketry, Advanced Ballistics, Particle Physics, Satellites, and Nanotechnology. All Space Shuttle parts require an aluminum mine connected to the empire.

A domination victory is achieved by sacking every other agent's capital city. A capital city can be sacked by causing its health to fall to zero or below during a war.

A culture victory is achieved when an agent's tourism output has eclipsed the cumulative culture output of each other agent in the game individually. Tourism is generated through various means, such as certain buildings, Great Works, Great People, and more\footnote{For more information, see \url{https://trevormcinroe.github.io/terra_nova_environment\#gps-mech}}. A given agent can accelerate its tourism pacing against another agent by establishing trade routes with the other agent or by exerting enough religious pressure on the other agent's cities such that they convert. 

A diplomatic victory is achieved by receiving 12 votes to be World Leader during a World Congress meeting. World Congress meetings begin after any agent in the game has met every other agent and researched the Printing Press. The World Congress convenes every 30 turns and votes can be gained via two means. First, agents can earn special delegates to the World Congress by constructing certain World Wonders\footnote{For information on World Wonders, see \url{https://trevormcinroe.github.io/terra_nova_environment\#buildings-mech}} or unlocking certain social policies\footnote{For information on social policies, see \url{https://trevormcinroe.github.io/terra_nova_environment\#culture-mech}}. Second, agents can convince their city-state allies to vote for them\footnote{For information on city-states, see \url{https://trevormcinroe.github.io/terra_nova_environment\#cs-mech}}.

The four victory conditions described above require distinct strategies to achieve because the game components in Terra Nova are highly specialized. For example, buildings and social policies that help agents produce enough science output to reach deep into the technology tree provide no direct benefits towards cultural output or exerting influence over city-states. Ergo, investing turns towards prerequisites for a science victory provides no direct progress toward a cultural or diplomatic victory.

\section{Previous Work in Civilization}\label{sec:previous-work}
\citet{amato-civ} learn to play Civilization IV by selecting one of four high-level policies that execute distinct playstyles via a pre-programmed low-level controller.~\citet{branavan-civ} learn to play Civilization II by combining MCTS and reasoning over the language in the game manual, but limit the game to a single victory condition. Recently,~\citet{civrealm} investigate language-guided control in Civilization II, but primarily focus on minigames and rely on a pre-trained language model to provide structured hints to the agent at each turn. In contrast to these previous works, we aim to provide researchers with access to an analogue of the full game of Civilization V.

\section{Formalizing Terra Nova as a Game}\label{sec:formalizing}
Game turns in Terra Nova are the cumulative effects of six agent turns. Terra Nova could be formalized under several different paradigms (e.g.,~\citep{openspiel,bram_paper}). Here, we view it as a turn-based partially observable stochastic game (POSG)~\citep{shapley_posg} parameterized by the tuple $\mathcal{G} = \langle \mathcal{N}, \mathcal{X}, \mathcal{A}, \mathcal{O}, \Omega, T, R, \rho, \mathcal{V}, \zeta \rangle$ where 

\begin{itemize}
    \item $\mathcal{N} = \{1, \dots, n\}$ is a finite set of $n$ agents\footnote{Full games of Terra Nova are always played between six agents.}.
    \item $\mathcal{X}$ is the (possibly infinite) set of environment states, 
which includes both discrete and continuous components, the ID of the currently-active agent, the current timestep, and game turn.
    \item $\mathcal{A}$ is a finite set of discrete actions.
    \item $\mathcal{O}$ is the (possibly infinite) set of observations, which includes discrete and continuous components.
    \item $\Omega: \mathcal{X} \rightarrow \mathcal{O}$ is a function that maps environment states to observations.
    \item $T: \mathcal{X} \times \mathcal{A} \times \mathcal{X} \rightarrow [0,1]$ is the transition kernel.
    \item $R: \mathcal{X} \times \mathcal{A} \times \mathcal{X} \rightarrow \mathbb{R}$ is the reward function.
    \item $\rho$ is the initial state distribution on $\mathcal{X}$.
    \item $\mathcal{V}$ is the finite set of victory types\footnote{Including ``no victory''.}.
    \item $\zeta : \mathcal{X} \rightarrow \mathcal{V}$ is the victory-status function that maps environment states to a victory type. 
\end{itemize}

For each agent $i \in \mathcal{N}$ and state $x \in \mathcal{X}$, let 
$\mathcal{A}^i(x) \subseteq \mathcal{A}$ denote the set of actions available to agent $i$ in state $x$. Here, we use superscripts to denote agent indices and subscripts to denote time indices. At the beginning of each game, an initial state is sampled $x_1 \sim \rho$. The first agent receives an observation $o^1_1 = \Omega(x_1)$ and selects an action $a^1_1 \in \mathcal{A}^1(x_1)$. The POSG transitions to the next state $x_2 \sim T(\; \cdot \; \vert \; x_1,a^1_1)$ and the first agent receives a reward $r^1_1 = R(x_1, a^1_1, x_2)$. This process repeats for each agent in $\mathcal{N}$ until a maximum number of game turns is reached or if $\zeta(x_t)$ returns any element in $\mathcal{V}$ other than ``no victory''. In Terra Nova, there are six POSG steps (one per player) per game turn.

We avoid defining the notion of agent, policy, or learning objective as agents in Terra Nova could be studied through a multitude of lenses. For example, an agent that maximizes returns in a given game might not satisfy any of the game's victory conditions (see \S\ref{sec:challenges}). This is because Terra Nova's reward function provides rewards for playing the game well (e.g., growing city population, building World Wonders, etc) instead of rewarding progress towards any one victory type. This is analogous to providing a chess-playing agent rewards for developing pieces and finding tactics as opposed to providing a reward only when the game is won or lost. Alternatively, one could choose to study agents in Terra Nova under a sparse reward function ($R_{\mathrm{sparse}}(x_t,a^i_t,x_{t+1}) := \mathds{1}\big[ \zeta(x_{t+1}) \neq \text{``no victory''} \big]$) to test agents' game-winning capabilities instead of game-playing capabilities. We highly encourage researchers to both pursue the goal of designing learning agents that can win Terra Nova consistently and to study agents in the environment in ways other than reward maximization machines.

\section{Software}\label{sec:software}

\begin{figure}[!t]
    \centering
    \includegraphics[width=0.95\linewidth]{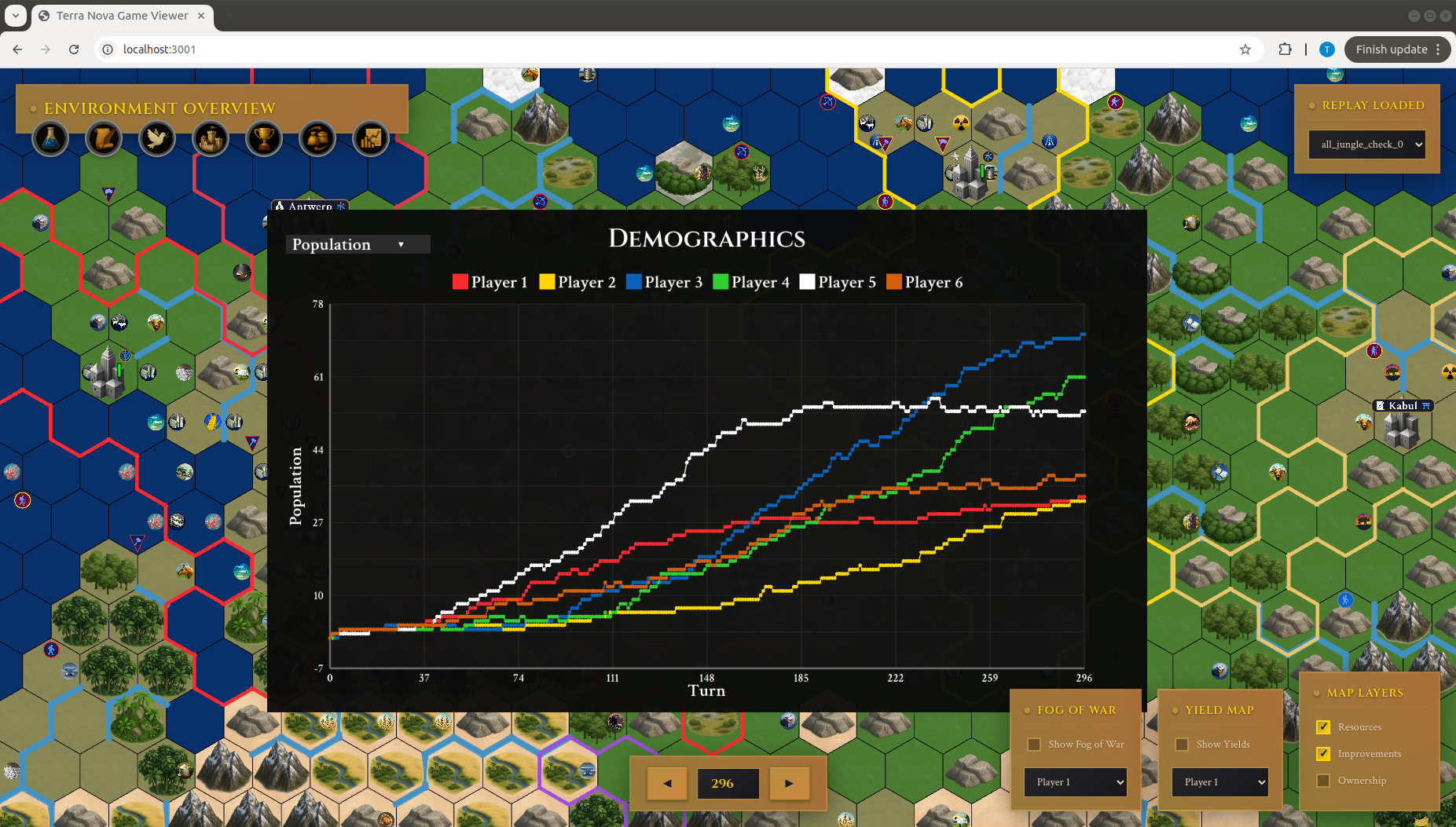}
    \caption{An example demographics screen in the Terra Nova Viewer. Displayed here is the total population in each agent's empire plotted over 296 game turns. Users can view many other statistics using the dropdown menu on the top-left of the Demographics screen. Additionally, the ``Environment Overview'' bar on the top-left of the Viewer contains buttons for many other information screens, and the map is zoomable and draggable, giving the user a complete view of the game.}
    \label{fig:pop-demos}
\end{figure}

The Terra Nova codebase comes with many utilities to help researchers. Below, we briefly highlight a few of the tools we provide for Terra Nova users. For information on each of the components below and more, see the project's website: \url{https://trevormcinroe.github.io/terra_nova}.

\textbf{Maps.} Terra Nova's initial release comes with 10k maps, which can be downloaded here: \url{https://huggingface.co/datasets/trevormcinroe/terra_nova_maps}. Maps are procedurally generated by the Terra Nova game engine; the generation process closely matches the map generation process used in LekMod\footnote{\url{https://www.marynkruithof.nl/the-lekmod-project/}}~\citep{lekmod}, a community-driven game mod specifically designed to balance Civilization V around free-for-all multiplayer games. The maps are designed to be balanced (in terms of resource distribution) across six equally-sized sections, one for each agent. Each section is given a ``regional resource'',  which is spawned many times within the section and rarely in other sections. In theory, this gives each player a monopoly over their regional resource, which can be advantageous in trade agreements. In addition to the regional resource, each section is populated with many other resources to create a large variety of fruitful city-settling locations. We note that an agent is not restricted to settling cities within their region. See Figure~\ref{fig:map-zoomout} for an example map.

\textbf{Distributed games.} The Terra Nova software will automatically distribute games across any XLA devices that are made visible. To accomplish this, Terra Nova leverages JAX's shard map utilities\footnote{For more information, see \url{https://docs.jax.dev/en/latest/notebooks/shard_map.html}}. Therefore, researchers can achieve higher throughput during training by scaling compute via parallel games.

\textbf{Recording and viewing games.} Terra Nova comes with utilities for recording and viewing games. These recordings are compressed and saved to disk. Recordings can be loaded into the Terra Nova Viewer, allowing the user to have full visibility of the game. This includes being able to see within cities, track every agent's progress in the technology and social policy trees, demographics over time, and more. See Figure~\ref{fig:pop-demos} for an example ``Population Demographics'' screen.

\textbf{Neural network.} Due to the complexity of the observation and action space, we provide researchers with a starting point on neural network architecture. In short, the starter architecture contains an encoder for each category of information in the observation (see the documentation on the observation space for more detail: \url{https://trevormcinroe.github.io/terra_nova_documentation\#observation-space}). The network also handles spatial information by scattering values onto a map (similar to~\citet{alphastar_unplugged}), and forces these representations through a bottleneck with cross attention between relevant components. The network ultimately has a learnable action head for each sub-action space (\S\ref{sec:challenges}). We hope that this starting point will accelerate research with Terra Nova.

\textbf{Environment API.} Terra Nova's API is designed to be gym-like. In the code block below, we provide an example for how a user might load a set of maps, build their distributed games simulator, and step the environment for each agent.

\begin{lstlisting}[language=Python]
import argparse
import os
import pickle
import jax

from sim.build import build_simulator


parser = argparse.ArgumentParser()
parser.add_argument("--seed", type=int)
parser.add_argument("--num_steps", type=int, default=300)
parser.add_argument("--map_folder", type=str)
parser.add_argument("--distributed_strategy", type=str)
args = parser.parse_args()

all_maps = os.listdir(args.map_folder)

games = []

for game in all_maps:
    if ".gamestate" not in game:
        continue
    with open(f"{args.map_folder}/{game}", "rb") as f:
        gamestate = pickle.load(f)
    games.append(gamestate)

(
  env_step_fn, games, obs_spaces, episode_metrics,
  players_turn_id, obs, GLOBAL_MESH, sharding
) = build_simulator(games, args.distributed_strategy, jax.random.PRNGKey(args.seed))

for game_step in range(args.num_steps):
    for agent_step in range(6):
        actions = ...
        (
          games, obs_spaces, episode_metrics, new_players_turn_id,
          next_obs, rewards, done_flags, selected_actions
        ) = env_step_fn(games, actions, obs_spaces, episode_metrics, players_turn_id)

        players_turn_id = new_players_turn_id
        obs = next_obs
\end{lstlisting}










\section*{Acknowledgments}
We would like to thank Joe Kelliher for their contributions of manually testing the game engine for bugs, Elle Miller for their thoughts on framing, Samuel Garcin, Kale-ab Tessera for their invaluable insights on what makes games like Civilization a strong intelligence testbed, David Abel for their always-valuable alternative perspective, and Sid Meier for providing us with countless hours of fun.




\bibliography{main}
\bibliographystyle{tmlr}


\end{document}